\documentclass[runningheads]{llncs}
\usepackage{graphicx}
\usepackage{amsmath,amssymb} 
\usepackage{color}
\usepackage[width=122mm,left=12mm,paperwidth=146mm,height=193mm,top=12mm,paperheight=217mm]{geometry}
\begin{document}
\pagestyle{headings}
\mainmatter
\def\ECCV18SubNumber{14}  

\title{Improving the Annotation of DeepFashion Images for Fine-grained Attribute Recognition} 

\titlerunning{Improving the Annotation of DeepFashion}

\authorrunning{R. Zakizadeh, M. Sasdelli, Y. Qian and E. Vazquez}

\author{Roshanak Zakizadeh, Michele Sasdelli, Yu Qian and Eduard Vazquez}
\institute{Cortexica Vision Systems, London, UK}

\maketitle

\begin{abstract}
DeepFashion is a widely used clothing dataset with 50 categories and more than overall 200k images where each image is annotated with fine-grained attributes. This dataset is often used for clothes recognition and although it provides comprehensive annotations, the attributes distribution is unbalanced and repetitive specially for training fine-grained attribute recognition models. In this work, we tailored DeepFashion for fine-grained attribute recognition task by focusing on each category separately. After selecting categories with sufficient number of images for training, we remove very scarce attributes and merge the duplicate ones in each category, then we clean the dataset based on the new list of attributes. We use a bilinear convolutional neural network with pairwise ranking loss function for multi-label fine-grained attribute recognition and show that the new annotations improve the results for such a task. The detailed annotations for each of the selected categories are provided for public use.

\end{abstract}

\section{Introduction}
Multi-label attribute recognition of an item (such as a single instance of a bird, a car or a dress) at a fine-grained level includes retrieving the detailed attributes which describe that item, for instance a dress can be labelled for its pattern (e.g. striped), length (e.g. maxi), fabric (e.g. chiffon), etc. There are very few \textit{publicly available} datasets which provide such detailed annotations including CUB200 Birds~\cite{wah2011caltech}, DeepFashion~\cite{liu2016deepfashion} and some face datasets such as LFW~\cite{learned2016labeled}.

DeepFashion dataset is mostly used for clothes recognition task, it contains over 200k images from 50 categories of clothes including Dress, Kimono, Shorts, etc. There are overall 1000 attributes describing images at a fine-grained level. However, not all of these categories, by themselves, contain enough images to train a fine-grained attribute recognition model. Further, not all of the 1000 attributes apply to images in every category. To this end, in this paper, we focus on nine relatively large categories of DeepFashion dataset and by removing the scare attributes and merging the visually similar ones will make these nine categories of clothes more suitable for the fine-grained attribute recognition task. We make the training, test and validation sets with their new list of annotations per category available for public use~\footnote{The nine categories with updated annotations are available for download \href{https://github.com/roshanaz/deepfashion_ninecat}{here}.}.
 
\section{DeepFashion for Fine-grained Attribute Recognition}

\begin{figure}
    \centering
    \includegraphics[trim={5cm 6.8cm 5cm 6.5cm},clip, width=\linewidth]{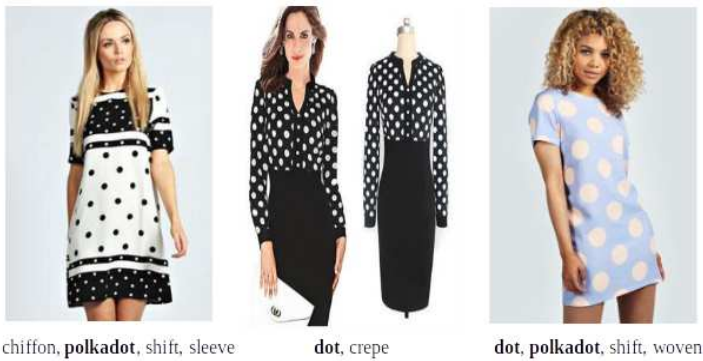}
    \caption{Examples of repetitive attributes in DeepFashion dataset}
    \label{fig:duplabels}
\end{figure}

The following steps explain the process of extending notations of DeepFashion dataset per category of clothes:

\begin{enumerate}
    \item \textbf{Fine-grained  categories with sufficient training  samples:} Each fine-grained  category in DeepFashion dataset contains different number of images. Out of the 50 categories 35 have only a few hundred images or less, among the remaining categories again six more contain around 5000 or less samples. Through empirical experiments with the convolutional neural network that we have used for fine-grained attribute recognition, we came to conclusion that at least 6000 samples are required for the loss to converge. After removing categories with not enough images, nine categories out of the original 50 remain.  These remaining categories and the number of images for each are: dress (50837), tee (24956), blouse (17095), shorts (13637), tank (10692), skirt (10568), cardigan (9050), sweater (8901) and top (7053).

    \item \textbf{Available attributes and enough attributes:}
    There are in total 1000 attributes annotated for DeepFashion dataset. However, not all of these attributes apply to all 50 fine-grained categories. So, we had to check within each category of clothing whether there are any samples available for each attribute. For instance, there are only 690 attributes assigned to the images in the \textit{cardigan} category. This means if the number of classes for fine-grained attribute recognition is set to the original 1000 attributes, for the 310 remaining attributes there are no samples available in the \textit{cardigan} category and this has to be noted during the design of the training model. Further, we need to consider the attribute imbalance problem, the ratio of the attributes for some fine-grained  categories is about 1:10000. To mitigate this problem, we set a threshold  for how few the distribution of samples can be for an attribute and discard attributes with the distribution of almost less than $2\%$ of the images within each category. Even after discarding such scarce attributes most categories are still unbalanced, but if we set the threshold  higher  than  $2\%$ there will be very few attributes left per category. Following this step a second pass is required  to remove images which now have no attributes. The final training size of the nine categories is as follows: dress (45869), tee(20216), blouse (15030), shorts (10641), skirt (9943), tank (8601), cardigan (7861), sweater (7273) and top (6083).
    
    \item \textbf{Merging duplicate annotations:} 
    After close investigation of the attributes we realized there are some annotated attributes which are very close in definition. This is especially true for the texture descriptive attributes such as striped, printed, dot, etc. An example of this can be seen in Figure~\ref{fig:duplabels} where the three dresses are annotated as polkadot, dot or even both (see the last image on the right), however all three dresses are visually recognized as the same pattern.  There are several examples of this kind in DeepFashion dataset. Overlapping attributes contributes to the imbalance of the dataset. We have improved the annotations by merging the visually similar attributes and have removed the duplicates. After removing the duplicate and scare attributes, the number of attributes per category is as follows: blouse (35), cardigan (36), sweater (31), tank (31), tee(23), top (29), shorts (31), skirt (32) and dress (35).
\end{enumerate}

\section{Experiments and Results}

\begin{figure}
    \centering
    \includegraphics[trim={1cm 9cm 3cm 5cm},clip, width=\linewidth]{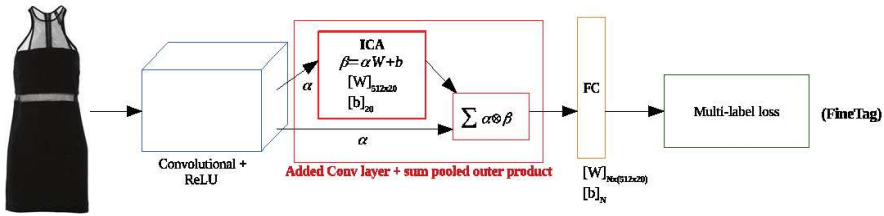}
    \caption{The bilinear network with pairwise ranking loss for extracting the attributes at fine-grained level in an image.}
    \label{fig:net}
\end{figure}

For fine-grained attribute recognition we chose the model shown in Figure~\ref{fig:net} (which we call FineTag). The model is a VGG16~\cite{simonyan2014very} based fully convolutional architecture for multi-label attribute recognition at fine-grained level. It has a bilinear-pool layer~\cite{lin2015bilinear} and uses a pairwise ranking loss function~\cite{li2017improving}. We use VGG16 pre-trained weights~\cite{deng2009imagenet,krizhevsky2012imagenet} on the ImageNet dataset~\cite{deng2009imagenet} only for initializing the convolutional layers. Then  we truncate the model at the last convolutional layer after the non-linearities and  generate $\beta$ by projecting a copy of the feature map ($\alpha$) into a 20 dimensional  ICA projection space~\cite{hyvarinen1999survey} (which is generated beforehand based on the feature maps from the same dataset). Then, the sum of the outer product of $\alpha$ and $\beta$ at each location is calculated which is passed through a fully connected layer.  The loss function used is the smooth pairwise ranking loss function proposed by~\cite{li2017improving}. The number of labels is different depending on the category, for instance there are 35 attributes for blouse category after merging the duplicates. The reason to choose FineTag architecture for our experiment is that it requires much less parameters to train compared to very deep networks like VGG16 but can produce results with the same precision score. Depending on the number of attributes this model is about $40\times$ smaller than VGG16 in terms of parameters.

We trained the model for each of the nine categories mentioned above twice: before and after merging the repetitive attributes. The model is built on Tensorflow framework and the experiments are run on an NVIDIA Tesla V100 GPU. For each category the model is trained in average for 30 epochs with the batch size of 40 using Adam optimizer~\cite{kingma2014adam} with the learning rate of 0.000001.




In Table~\ref{summary} for each category, we are reporting the results for the ranking-based average precision~\cite{furnkranz2008multilabel} and the weighted mean average precision~\cite{schutze2008introduction,zhao2015deep} weighting by the frequency of instances per label before and after removing the duplicates (denoted as AvgPrec\_b, AvgPrec\_a, wmap\_b and wmap\_a respectively) . We can see that there is significant improvements in the results just by merging the duplicated attributes similar to those in Figure~\ref{fig:duplabels}.

\setlength{\tabcolsep}{4pt}
\begin{table}
\begin{center}
\caption{Ranking-based average precision (AvgPrec) over all images and weighted mean average precision (wmap) before and after removing the duplicate attributes}
\label{summary}
\begin{tabular}{lllll}
\hline\noalign{\smallskip}
category    &    AvgPrec\_b   & AvgPrec\_a   &  wmap\_b     & wmap\_a  \\
\noalign{\smallskip}
\hline
\noalign{\smallskip}
blouse            & 0.49  & \textbf{0.51}  & 0.31 &  \textbf{0.33}  \\ 
cardigan          & 0.48	& \textbf{0.49}  & 0.27 &  \textbf{0.28}  \\
sweater            & 0.50 & \textbf{0.54}      & 0.31 &  \textbf{0.35}  \\ 
tank            & 0.50 & \textbf{0.53}      & 0.30 &   \textbf{0.33} \\ 
tee &	0.60	 & \textbf{0.63}      & 0.37 &  \textbf{0.41}  \\ 
top &	0.54				 & \textbf{0.58}      & 0.34 &  \textbf{0.38}  \\ 
shorts &	0.56		 & \textbf{0.57}      & 0.38 &  \textbf{0.39}  \\ 
skirt &	0.65				 & \textbf{0.66}      & 0.48 &  \textbf{0.49}  \\ 
dress &			0.59		 & \textbf{0.61}      & 0.40 &  \textbf{0.42}  \\ 
\hline
\end{tabular}
\end{center}
\end{table}
\setlength{\tabcolsep}{1.4pt}

Table~\ref{attributes} shows the weighted mean average precision score for two selected pairs of duplicate attributes per category (depending on the category the number of duplicate pairs vary, here we have chosen two pairs per category for demonstration). The categories are grouped based on the common attributes (\textit{shorts} category has only one pair of repetitive attributes) and the merged attribute for each pair is indicated by adding an \textbf{\_m} and in bold. We can see that the results per attribute improve significantly by merging the duplicate annotations in the dataset. Further, before merging the duplicates one of the attributes in the duplicate pair is often learned and extracted very poorly by the network which results in a low mean average precision score over the whole dataset. For instance, the \textit{printed} label for the \textit{tee} category in Table~\ref{attributes} is recognized by the network with a very low score of $0.06$ and is improved significantly (to \textit{printed\_m} with $0.54$ precision score) after being merged with the \textit{print} attribute. It is important to notice that per attribute precision scores are significantly important for retrieval applications where the query has a specific attribute that we are interested in. For instance, in case of querying a \textit{striped} dress we can see that the improved score of \textbf{striped\_m} could enhance the search results significantly.

\setlength{\tabcolsep}{4pt}
\begin{table}
\begin{center}
\caption{Weighted mean average precision (wmap) over all labels before and after merging the duplicate attributes}
\label{attributes}
\begin{tabular}{lllllll}

\hline \noalign{\smallskip}
category & \multicolumn{6}{c}{attributes}\\
 \noalign{\smallskip}
\hline
   &  print    & printed & \textbf{printed\_m} & stripe   & striped & \textbf{striped\_m}\\
tee     &  0.49  & 0.06	& \textbf{0.54}  & 0.40 & 0.42 & \textbf{0.75}  \\

top     &  0.50  & 0.06	& \textbf{0.54}  & 0.24 & 0.45 & \textbf{0.64}  \\

dress     &  0.57  & 0.10	& \textbf{0.62}  & 0.33 & 0.42 & \textbf{0.71}  \\

shorts     &  0.60  & 0.12	& \textbf{0.65}  & - & - & -  \\

\hline 
& & & & & & \\
    &  crop    & cropped & \textbf{cropped\_m} & stripe   & striped & \textbf{striped\_m}\\
sweater     &  0.10  & 0.31	& \textbf{0.41}  & 0.33 & 0.55 & \textbf{0.78}  \\

tank     &  0.30  & 0.12	& \textbf{0.42}  & 0.28 & 0.52 & \textbf{0.69}  \\

\hline
& & & & & & \\
    &  dot    & polkadot & \textbf{polkadot\_m} & print   & printed & \textbf{printed\_m}\\
blouse            & 0.60  & 0.61 & \textbf{0.65}  & 0.52 & 0.09 &  \textbf{0.59}  \\ 

\hline
& & & & & & \\
    &  dot    & polkadot & \textbf{polkadot\_m} & stripe   & striped & \textbf{striped\_m}\\
skirt     &  0.39  & 0.53	& \textbf{0.53}  & 0.27 & 0.46 & \textbf{0.64}  \\

\hline
& & & & & & \\
    &  fringe    & fringed & \textbf{fringed\_m} & stripe   & striped & \textbf{striped\_m}\\
cardigan     &  0.18  & 0.18	& \textbf{0.32}  & 0.16 & 0.55 & \textbf{0.62}  \\

\hline
\end{tabular}
\end{center}
\end{table}
\setlength{\tabcolsep}{1.4pt}

\section{Conclusions}
In this paper, we extracted nine categories of clothes from the DeepFashion dataset which provide sufficient samples and comprehensive annotations for fine-grained attribute recognition. Further, we showed merging duplicate attributes for DeepFashion improves the attribute recognition results over the samples and per attribute. This is mainly because duplicate attributes contribute to poor annotation of the images and one of the two repetitive attributes is always under-sampled which results in poor overall results.

\bibliographystyle{splncs}
\bibliography{egbib}

\begin{thebibliography}{10}

\bibitem{wah2011caltech}
Wah, C., Branson, S., Welinder, P., Perona, P., Belongie, S.:
\newblock The caltech-ucsd birds-200-2011 dataset.
\newblock (2011)

\bibitem{liu2016deepfashion}
Liu, Z., Luo, P., Qiu, S., Wang, X., Tang, X.:
\newblock Deepfashion: Powering robust clothes recognition and retrieval with
  rich annotations.
\newblock In: Proceedings of the IEEE conference on computer vision and pattern
  recognition. (2016)  1096--1104

\bibitem{learned2016labeled}
Learned-Miller, E., Huang, G.B., RoyChowdhury, A., Li, H., Hua, G.:
\newblock Labeled faces in the wild: A survey.
\newblock In: Advances in face detection and facial image analysis.
\newblock Springer (2016)  189--248

\bibitem{simonyan2014very}
Simonyan, K., Zisserman, A.:
\newblock Very deep convolutional networks for large-scale image recognition.
\newblock arXiv preprint arXiv:1409.1556 (2014)

\bibitem{lin2015bilinear}
Lin, T.Y., RoyChowdhury, A., Maji, S.:
\newblock Bilinear cnn models for fine-grained visual recognition.
\newblock In: Proceedings of the IEEE International Conference on Computer
  Vision. (2015)  1449--1457

\bibitem{li2017improving}
Li, Y., Song, Y., Luo, J.:
\newblock Improving pairwise ranking for multi-label image classification.
\newblock (2017)

\bibitem{deng2009imagenet}
Deng, J., Dong, W., Socher, R., Li, L.J., Li, K., Fei-Fei, L.:
\newblock Imagenet: A large-scale hierarchical image database.
\newblock In: Proceedings of the IEEE Conference on Computer Vision and Pattern
  Recognition (CVPR), IEEE (2009)  248--255

\bibitem{krizhevsky2012imagenet}
Krizhevsky, A., Sutskever, I., Hinton, G.E.:
\newblock Imagenet classification with deep convolutional neural networks.
\newblock In: Advances in neural information processing systems. (2012)
  1097--1105

\bibitem{hyvarinen1999survey}
Hyv{\"a}rinen, A.:
\newblock Survey on independent component analysis.
\newblock (1999)

\bibitem{kingma2014adam}
Kingma, D.P., Ba, J.:
\newblock Adam: A method for stochastic optimization.
\newblock arXiv preprint arXiv:1412.6980 (2014)

\bibitem{furnkranz2008multilabel}
F{\"u}rnkranz, J., H{\"u}llermeier, E., Menc{\'\i}a, E.L., Brinker, K.:
\newblock Multilabel classification via calibrated label ranking.
\newblock Machine learning \textbf{73}(2) (2008)  133--153

\bibitem{schutze2008introduction}
Sch{\"u}tze, H., Manning, C.D., Raghavan, P.:
\newblock Introduction to information retrieval. Volume~39.
\newblock Cambridge University Press (2008)

\bibitem{zhao2015deep}
Zhao, F., Huang, Y., Wang, L., Tan, T.:
\newblock Deep semantic ranking based hashing for multi-label image retrieval.
\newblock In: Proceedings of the IEEE Conference on Computer Vision and Pattern
  Recognition (CVPR), IEEE (2015)  1556--1564

\end{thebibliography}
\end{document}